\documentclass{ecai} 
\usepackage{latexsym}
\usepackage{amssymb}
\usepackage{amsmath}
\usepackage{amsthm}
\usepackage{booktabs}
\usepackage{enumitem}
\usepackage{graphicx}
\usepackage{color}
\usepackage{subcaption}
\usepackage{hyperref}
\begin{document}

%%%%%%%%%%%%%%%%%%%%%%%%%%%%%%%%%%%%%%%%%%%%%%%%%%%%%%%%%%%%%%%%%%%%%%%%

\begin{frontmatter}

%%% Use this command to specify your submission number.
%%% In doubleblind mode, it will be printed on the first page.

\paperid{2672} 

%%% Use this command to specify the title of your paper.

\title{A Neural Rewriting System \\ to Solve Algorithmic Problems}

%%% Use this combinations of commands to specify all authors of your 
%%% paper. Use \fnms{} and \snm{} to indicate everyone's first names 
%%% and surname. This will help the publisher with indexing the 
%%% proceedings. Please use a reasonable approximation in case your 
%%% name does not neatly split into "first names" and "surname".
%%% Specifying your ORCID digital identifier is optional. 
%%% Use the \thanks{} command to indicate one or more corresponding 
%%% authors and their email address(es). If so desired, you can specify
%%% author contributions using the \footnote{} command.

\author[A]{\fnms{Flavio}~\snm{Petruzzellis}\thanks{Corresponding Author. Email: flavio.petruzzellis@phd.unipd.it}}
\author[A,B]{\fnms{Alberto}~\snm{Testolin}}
\author[A]{\fnms{Alessandro}~\snm{Sperduti}} 

\address[A]{Department of Mathematics, University of Padova, Padova, Italy}
\address[B]{Department of General Psychology, University of Padova, Padova, Italy}

%%% Use this environment to include an abstract of your paper.

\begin{abstract}
Modern neural network architectures still struggle to learn algorithmic procedures that require to systematically apply compositional rules to solve out-of-distribution problem instances. In this work, we focus on formula simplification problems, a class of synthetic benchmarks used to study the systematic generalization capabilities of neural architectures. We propose a modular architecture designed to learn a general procedure for solving nested mathematical formulas by only relying on a minimal set of training examples. Inspired by rewriting systems, a classic framework in symbolic artificial intelligence, we include in the architecture three specialized and interacting modules: the Selector, trained to identify solvable sub-expressions; the Solver, mapping sub-expressions to their values; and the Combiner, replacing sub-expressions in the original formula with the solution provided by the Solver. We benchmark our system against the Neural Data Router, a recent model specialized for systematic generalization, and a state-of-the-art large language model (GPT-4) probed with advanced prompting strategies. We demonstrate that our approach achieves a higher degree of out-of-distribution generalization compared to these alternative approaches on three different types of formula simplification problems, and we discuss its limitations by analyzing its failures.
\end{abstract}

\end{frontmatter}

%%%%%%%%%%%%%%%%%%%%%%%%%%%%%%%%%%%%%%%%%%%%%%%%%%%%%%%%%%%%%%%%%%%%%%%%

\section{Introduction}
Whether neural networks can learn to perform systematic reasoning is a long-standing question in artificial intelligence and cognitive science \cite{Fodor1988-am}, and the recent success of deep learning on challenging reasoning tasks has reinvigorated this debate \cite{hupkes2020compositionality}. However, despite some promising achievements \cite{Lake2023-qb}, there is general consensus that even state-of-the-art transformer architectures and large language models still lack systematic and compositional generalization skills \cite{DBLP:conf/emnlp/CsordasIS22,dziri2023faith,press-etal-2023-measuring}.

In this work, we investigate this issue by considering a type of systematic generalization benchmark that requires simplifying symbolic formulas to a minimal form.
Specifically, we consider the problem of simplifying nested formulas of operations on lists of integers \cite{DBLP:conf/naacl/NangiaB18}, as well as arithmetic and algebraic expressions.
In principle these problems could be tackled by automatic symbolic solvers, but here we consider them as benchmarks to assess the reasoning capabilities of trained neural networks under controlled settings \cite{DBLP:conf/iclr/CsordasIS22, DBLP:conf/icml/LakeB18}.
%Despite its apparent simplicity (these problems could be effectively tackled by automatic symbolic solvers), \emph{learning} to solve this kind of problems is extremely challenging, as it implies the discovery of algorithmic procedures that generalize to formulas with different nesting depths and number of operands.

We formalize formula simplification problems and propose a simple solution method inspired by rewriting systems rooted in the tradition of symbolic artificial intelligence \cite{dershowitz1990rewrite}.
% We consider this as a reference method in the analysis of the performance of neural models on these tasks.
%and by the evidence that learning algebra retrains our visual system to perceive the hierarchical structure of symbolic expressions \cite{marghetis2016mastering}, 
We implement this solution method as a modular neural architecture called the Neural Rewriting System, which represents a case study of a neural architecture in which the roles and interaction mechanisms between neural modules are clearly defined, allowing greater interpretability of the resulting system.
We exploit our framework to further evaluate two neural architectures that have been recently shown able to tackle reasoning tasks: the Neural Data Router \cite{DBLP:conf/iclr/CsordasIS22}, a transformer encoder trainable end-to-end to solve algorithmic problems, and GPT-4 \cite{openai2023gpt4} probed using self-consistency zero-shot Chain-of-Thought \cite{DBLP:conf/iclr/0002WSLCNCZ23}, which is an advanced prompting method that improves the performance of large language models on reasoning tasks.

Through extensive experiments and analyses, we show that the proposed modular architecture achieves greater accuracy compared to advanced alternative approaches, especially when solving deeply nested formulas.
Furthermore, we show that our approach allows us to understand in detail what parts of a systematic generalization problem could represent the greatest obstacles for neural architectures.
Indeed, when analyzing the causes of failure of the Neural Rewriting System, we find that accurately selecting the portion to be simplified in very long formulas is the most challenging task for the architecture, shedding light on the limitations of current approaches to deal with length generalization in algorithmic tasks with transformers.

\section{Related works}
\textbf{Systematic Generalization}
A consistent stream of research investigates the extent to which neural networks can learn to reason in a systematic and compositional way \cite{hupkes2020compositionality,testolin2023can}.
Several benchmarks have been proposed to measure this capacity: among the most popular, ListOps \cite{DBLP:conf/naacl/NangiaB18} and CTL
\cite{DBLP:journals/corr/abs-1802-06467} can be seen as instances of formula simplification problems, in which the learning system is tasked to produce the evaluation of a formula which typically requires to compute intermediate results, either implicitly or explicitly.
Other formula simplification benchmarks include derivation and integration problems \cite{Lample2019DeepLF} or polynomial simplification \cite{agarwal2021analyzing}.
Modern neural architectures achieve different degrees of systematic compositional generalization
\cite{DBLP:conf/iclr/CsordasIS22,Lake2023-qb,DBLP:conf/acl/OntanonAFC22}.
%One such models is the Neural Module Network (NMN) \cite{Andreas2015NeuralMN}, whose systematic generalization properties on a visual question answering task were found to be highly dependant on the specific layout of the modules which proved to be hard to learn from data \cite{DBLP:conf/iclr/BahdanauMNNVC19}.
Interestingly, recent work has shown the emergence of specialized modules in Mixture of Experts architectures \cite{DBLP:conf/nips/MittalBL22}, suggesting that modular architectures can outperform monolithic ones in both in- and out-of-distribution testing conditions.% highly specialized modules do not emerge from standard end-to-end training without ad-hoc learning biases.

\textbf{Rewriting systems and neural networks}
The idea of implementing neural architectures inspired by symbolic rewriting systems was initially explored in unsupervised learning settings, where individual network weights represented tokens to be rewritten \cite{icnc09}. Others implemented a rewriting system for algebraic problems using custom feature engineering methods and a feed-forward network \cite{Cai2018-xx}, while in \cite{DBLP:conf/nips/ChenT19} the authors propose a reinforcement learning-based system that can learn a general rewriting mechanism by first choosing a region to simplify and then an appropriate rewriting rule.

\textbf{LLMs} 
Research on prompt engineering techniques has shown that appropriate prompting methods can boost the reasoning capabilities of LLMs \cite{DBLP:conf/iclr/0002WSLCNCZ23,DBLP:conf/nips/Wei0SBIXCLZ22}.
In particular, Chain-of-Though prompting improves LLMs' accuracy on reasoning tasks by leveraging their auto-regressive nature: the models are elicited to give the final result using multi-step reasoning chains, which enable to process contextual information more effectively.
Among the reasoning tasks on which Large Language Models often exhibit impressive problem-solving capabilities we find `symbolic reasoning tasks'.
Some examples are coin flip, last letter concatenation \cite{DBLP:conf/nips/Wei0SBIXCLZ22} and boolean variable assignment \cite{DBLP:conf/nips/AnilWALMRSGDN22}.
These tasks are remarkably similar in spirit to those we consider in this work, since problem instances are synthetically generated and can be solved by applying simple algorithms.
% However, LLMs still struggle with problems that require systematic generalization \cite{press-etal-2023-measuring}.

\section{Formula simplification problems}
\label{sec:fsp}
Many systematic generalization benchmarks require to evaluate a possibly very complex formula by replacing expressions with elementary operands that have an equivalent value. 
We call such problems `formula simplification problems'.
For example, in ListOps an operation on a list of integers can be replaced with a single integer that is the result of the operation.
In order to use formula simplification problems as a benchmark to assess the systematic generalization abilities of neural architectures, we outline a formal framework that includes the class of problems we consider in this work. 
%and which we will use to describe define more formally and describe their intuitive solution process in terms of this formal definition.

\begin{figure}[t]
    \centering
    \includegraphics[width=0.5\textwidth]{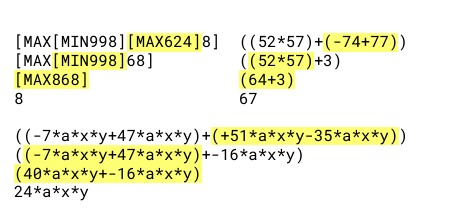}
    \caption{Examples of solution of ListOps, arithmetic and algebraic formulas. Formulas $f$ are reduced to atomic values $e$ by iteratively solving leaf formulas $f^L$, highlighted in yellow.\vspace{0.8cm}}
    \label{fig:solution-example}
\end{figure}

In formula simplification problems, we can see formulas $f \in F$ as entities that are composed of two semantically distinct elements: operators $o \in O$ and arguments $a \in A$. \footnote{We also sometimes refer to arguments as operands. We prefer to use the term arguments here to avoid confusion with operators $O$.}
Arguments can be either atomic elements $e \in E$, such as integers, which are also the final values of any formula, or other formulas:
therefore, in general, formulas have a recursive structure (see Figure \ref{fig:solution-example}).
Furthermore, since in these problems each formula has always one and only one final value, any formula has the generic compact form $f = o(a_1,...,a_n) = e$, where \mbox{$o\in O$, $e \in E$} and \mbox{$a_j \in F \cup E, ~\forall j \in [1, n]$.}
Finally, we can define leaf formulas \mbox{$F^L \subset F$} as the subset of formulas whose arguments are all atomic elements: \mbox{$f^L=o(a_1,...,a_n) \;\ \mathrm{s.t.}\;\  a_k \in E, ~\forall k \in [1,n]$}.

The simplest algorithm that can be applied to solve formula simplification problems is to solve leaf formulas one by one until the formula is completely reduced to an atomic element $e$.
Therefore, for any problem, we can define a set of rewriting rules $r \in R$ which map leaf formulas to their values: $r: F^L \rightarrow E, ~\forall r \in R$.
Taking into account the problems we consider in this work, we can further characterize the rewriting rules in $R$. First, the rules apply to contiguous sub-sequences of symbols.
Second, each rewriting rule is contractive; that is, it maps sub-sequences to shorter ones.
Third, the rewriting rules have mutually exclusive  applicability scope, i.e. given a leaf formula,  one and only one rule can be applied to reduce it.
Fourth,
several leaf formulas could be present in a formula at once, but the order in which rewriting rules are applied on them is irrelevant.

The notation introduced above allows us to outline a simple algorithm to solve nested formulas, consisting of the iterative application of the composition of three functions:
% According to the formalization outlined above, any algorithmic task which falls into this class of problems can be solved by iterative rewriting the initial formula applying a finite sequence of valid rewriting rules until the formula is completely reduced to an atomic element $e$.
%In general, if at any point in the iterative process there is no valid rule that can be applied to a leaf formula, the computation ends.
% More precisely, we can identify three sub-tasks which that should be performed to solve these problems: 
$sel: F \rightarrow F^L$, that maps a formula to a leaf formula appearing in it; $sol: F^L  \rightarrow E$ that solves the leaf formula, i.e. applies an appropriate rewriting rule to map it to an equivalent atomic element; $com: F \times F^L \times E \rightarrow F$, that combines the initial formula $f$ and the solution of the leaf formula $f^L$ by replacing $f^L$ with $sol(f^L)$ in the initial formula.
Notice that the function $sel$ can be applied again on the output of the function $com$ which is a valid formula, thus enabling the implementation of an iterative solution procedure.

Throughout this work, we will use this simple algorithm as a reference to compare the way different models solve formula simplification problems and to critically evaluate their results. Furthermore, we propose a neural implementation of this algorithm in the form of a modular neural architecture called the Neural Rewriting System.

% analyze its generalization capabilities on the whole formula simplification problem and on each sub-task, in order to understand the benefits and limitations of a modular approach to systematic generalization.

%In the next section, we describe an integrated modular neural architecture where each module tries to solve one of the three sub-tasks described above.

\section{Neural Rewriting System}
\label{sec:nrs}

\begin{figure}[t]
    \centering
    \includegraphics[width=0.5\textwidth]{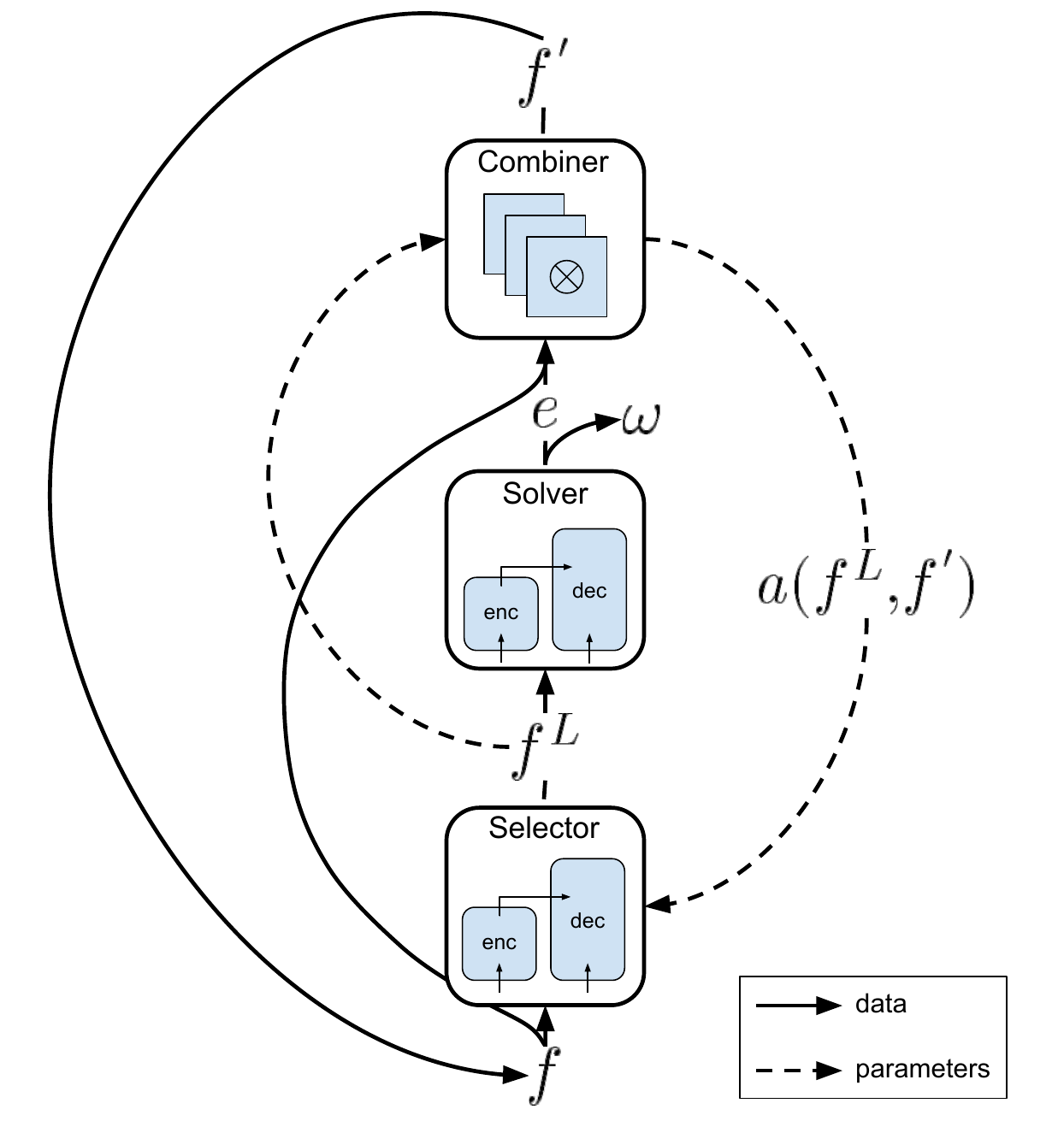}
    \caption{Schematic representation of the Neural Rewriting System.\vspace{0.8cm}}
    \label{fig:architecture}
\end{figure}

In this section, we describe the proposed Neural Rewriting System, a neural architecture composed of three integrated modules, called Selector, Solver, and Combiner, each being the neural implementation of one of the functions that can be used to solve formula simplification problems according to the algorithm described in Section \ref{sec:fsp}.
The modules are trained independently of each other and interact at test time to solve nested formulas iteratively.
The model is designed to achieve strong out-of-distribution generalization, i.e. to solve very deep formulas while being trained on a set of much shallower formulas.
We describe the composition of the datasets used in the development of the model in Section \ref{exp-nrs}.
A schematic representation of the system is shown in Figure \ref{fig:architecture}.
% Two out of the three modules are trained independently from each other, and the three modules interact to solve a given formula.

\label{subpar:sel}
\subsection{Selector Module}

The Selector module implements the function $sel: F \rightarrow F^L$, i.e. it is trained to map a formula to a leaf formula appearing therein.
In analogy to what happens in humans when they deploy object-based attention to locate algebraic sub-expressions that can be simplified \cite{marghetis2016mastering}, the Selector is trained to identify the last leaf formula\footnote{To enable generalization on formulas with more arguments than seen during training, if the identified leaf formula has more than two arguments, we train the Selector to output the smallest fraction of the leaf formula which can be simplified. This corresponds to the operator and the two first arguments appearing in the leaf formula.} occurring in the input formula on which a rewriting rule can be applied.
When training the Selector, we assume that it will always receive syntactically correct formulas, i.e. we use teacher forcing.
% The identified leaf formula is then given in input to the Solver.
% At inference time, the output of the Selector is also used to dynamically set the filters of a Convolutional Neural Network that is a component of the Combiner.
% Therefore, this module can also be seen as a HyperNetwork \cite{DBLP:conf/iclr/HaDL17} which sets the weights of the Combiner with discrete values.

We frame the problem as a sequence-to-sequence task and use a variant of the transformer encoder-decoder \cite{DBLP:conf/nips/VaswaniSPUJGKP17} to implement the Selector.
We modify the vanilla transformer in two ways. % which follow from assumptions that apply to the formula simplification problems we consider.
First, given the class of problems we consider in this work, we can assume that the problem the Selector needs to solve has a local nature: the leaf formula to be simplified is always located in a limited region of the input sequence.
We therefore mask all entries in the self-attention matrix of the encoder but the ones around the main diagonal (i.e., we make them $-\mathrm{inf}$).
The diagonal window is $2k+1$ tokens wide, where $k$ is a hyperparameter. 
\textcolor{black}{We also conducted preliminary experiments showing that models with vanilla self-attention achieved worse out-of-distribution generalization.}
Second, to enable generalization to out-of-distribution problem instances, we require the Selector to identify a leaf formula independently of the length of the input sequence, provided that it is syntactically correct.
Recent research on length generalization in transformers provided evidence that their ability to generalize on longer samples can be influenced by the choice of positional encodings, especially when, at test time, these fall out of the range observed during training \cite{DBLP:conf/emnlp/CsordasIS21,DBLP:conf/acl/RuossDGGCBLV23}. 
We thus use Label-based Positional Encodings \cite{DBLP:journals/corr/abs-2210-00400} to enable the Selector to identify leaf formulas in very long sequences.
The positional information of an input sequence of $L$ tokens is thus encoded in the following way: given a sequence of $N$ sinusoidal positional encodings, where $N$ is a large number that represents the maximum expected length of an input, $L$ integers are sampled in the interval $[0, N-1]$ and then sorted.
The encodings found in the positions corresponding to the sampled integers are then summed to the embeddings of the tokens in the input sequence.

\textbf{Multi-output generation} Label positional encodings introduce randomness in the representation of input sequences.
We have observed experimentally that to mitigate the effect of randomness it can be useful to repeat the auto-regressive generation of transformer outputs. 
After sampling several output sequences from the probability distribution derived from the decoder's outputs, we choose the best one considering both a measure of confidence of the Selector and a measure of input-output agreement computed by the Combiner.

Given the specialized purpose of the Selector, we can see each output of the module as a candidate leaf formula $\hat f^L$.
We generate any token $\hat f_i^L$ in an output sequence $\hat f^L$ by sampling from the probability distribution obtained applying the $\mathrm{softmax}$ function to the logits produced by the final fully-connected layer of the decoder.
We do not use any temperature parameter when sampling the output tokens.
For any input formula $f$, we repeat the stochastic generation process $M$ times, thus generating a sequence of candidate leaf formulas \mbox{$\hat F^L = \langle \hat{f}^{L,1},...,\hat{f}^{L,M}\rangle$}.
We define the confidence of the Selector on any $\hat f^{L,j}, 1 \leq j \leq M$ as the joint probability of sampling its tokens: $c(\hat f^{L,j})=\prod_{i=1}^{N} p_i^j$, where $N$ is the number of tokens in $\hat f^{L,j}$, and $p_i^j$ is the probability to sample token $\hat f_i^L$ in $\hat f^{L,j}$.
We also define an agreement score $a(\hat f^{L,j},f) \in [0,1]$ which gives information about the fraction of $\hat f^{L,j}$ which is exactly present in the input formula $f$.
This measure is computed by the Combiner and thus it is formally defined in Section \ref{subsec:com}.
We then select the final output $f^L$ of the Selector as the one with the highest Selector confidence which has an agreement score equal to $1$ --- that is, it matches the input sequence exactly.
More formally, \mbox{$f^L = \hat f^{L,j} \in \hat F^L \;$} $\mathrm{s.t.}\;\ $ \mbox{$c(\hat f^{L,j}) \geq c(\hat f^{L,k}) ~\forall j,k \in [1, M] \land a(\hat f^{L,j}, f) = 1$.}

\textbf{Dynamic windowing} We implement a dynamic windowing mechanism on longer input sequences that allows us to increase the model's generalization capacity on complex problem instances.
The core idea behind this mechanism is to repeat the process of selecting a leaf formula several times, changing each time the \textit{window} of the input formula that the Selector observes, and then relying on the confidence $c(\hat f^{L})$ to pick the best output.
We apply this mechanism on top of multi-output generation by modifying its behavior for sequences longer than a given threshold $T$.
Given an input formula $f$, if $|f|<T$ the computation is executed as described before.
Otherwise, we generate $M$ copies of the input $\langle f^{(1)},...,f^{(M)}\rangle$, whose lengths
%corresponding to the $M$ candidate leaf formulas $\langle \hat f^{L,1},..., \hat f^{L,M} \rangle$ generated by the Selector.
%The length of each $f^{(k)}$ 
will be reduced by applying a window function $w$.
Considering any input $f$ as a sequence $f_1,...,f_N$ of $N$ tokens, we define the window function $w(f,k)=f_{k+1},...,f_N$ which reduces the length of the input by giving as output its last $k$ tokens.
Since the Selector is trained to output the last leaf formula appearing in the input, the window function reduces the input length starting from the first tokens.
We divide the sequence of copies of the input $\langle f^{(1)},...,f^{(M)}\rangle$ into 20 groups $F^{(1)},...,F^{(20)}$ of equal size.
Intuitively, in each group the length of the input is reduced by a different percentage of tokens.
More formally, the window function will be parameterized by \mbox{$k=\mathrm{floor}(|f^{(i)}| \cdot \frac{j}{20}) ~\forall f^{(i)} \in F^{(j)}, ~\forall j \in [1, 20]$}.
We then pick the final leaf formula using the confidence and agreement scores, as described in the previous paragraph.
This ensures that the model can observe the whole input sequence and select a leaf expression in the part of the input where it can identify one with more confidence.

\subsection{Solver Module}
Unlike classical rewriting systems that use symbolic rule dictionaries, we learn rewriting rules directly from data by implementing the function $sol: F^L \rightarrow E$ with a trainable neural network. 
Given a leaf formula $f^L$ by the Selector, the Solver is trained to produce the equivalent reduction $e$ according to a valid rewriting rule.
When training the Solver, we assume that the Selector always generates perfect outputs (i.e., we use teacher forcing).
The Solver also learns to recognize the computation's termination state, signaling when such a state is reached. 
It maps atomic elements, representing a formula's final value, to the special symbol $\omega$, indicating the end of computation. 
Similar to the Selector, we frame the task as a sequence-to-sequence problem.
We use a vanilla transformer encoder-decoder for the Solver since it learns simple input-output mappings.

% The solution produced by the Solver will then be replaced in the original expression by the Combiner, thus producing a simplified version of the original formula.

\label{subsec:com}
\subsection{Combiner Module}
The last module in the architecture is the Combiner, a neural implementation of the function $com: F \times F^L \times E \rightarrow F$. Its purpose is thus to produce a simplified version of the original formula, given the formula itself $f$, the leaf formula $f^L$ identified by the Selector, and its reduction $e$ computed by the Solver.

In order to carry out its task, the first operation that the Combiner must perform is finding the position in $f$ where the leaf formula $f^L$ appears.
%(indeed, the Selector does not produce any information on the position of the leaf formula). 
We notice that the convolution is a suitable operation to detect which portion of an input sequence has the highest match with another sequence used as a filter, so we implement this operation using a 2D Convolutional Neural Network (CNN) whose filters are set dynamically at execution time using the output of the Selector, rather than being learned with backpropagation.
%\cite{lecun}

More precisely, we represent both the input sequence $f$ and the leaf formula $f^L$ as sequences of 1-hot vectors over the same vocabulary.
Since the leaf formulas found for different sequences in a batch can have different lengths, we pad each one with zeros to prevent the padding to match in the input.
Then, we set the filter of the 2D CNN to the 1-hot representation of $f^L$.
We refer to the CNN parameterized in this way as $\mathrm{CNN}_{f^L}$.
Doing so allows us to obtain from the output of the convolution both information on the location of the best match of $f^L$ in $f$ and on the number of tokens in $f^L$ that match $f$ exactly in some point.
Indeed, we can compute the location of the best match as $\mathrm{pos}(f^L, f)=\mathrm{argmax} ~\mathrm{CNN}_{f^L}(f)$.
Furthermore, we can compute the agreement score $a(f^L,f)=\frac{\mathrm{max} ~\mathrm{CNN}_{f^L}(f)}{|f^L|}$, where $|f^L|$ is the number of tokens in $f^L$.
Dividing by $|f^L|$ makes the score normalized, which allows us to compare the agreement scores of leaf formulas with different lengths.
Indeed, as described in Section \ref{subpar:sel}, the Selector uses this score for multi-output generation to discard the outputs that do not have an exact match in the input formula.
Notice that in this case the CNN is parameterized using candidate leaf formulas $\hat f^L_j$ whose accuracy scores with $f$ are compared.
If there is no Selector output such that $a(\hat f^L_j, f)=1$, the computation on the input sequence $f$ is stopped, and this is considered a failure of the model.

After finding the position of the leaf expression in $f$, the Combiner replaces $f^L$ with $e$ in $f$, to compute the simplified version of the formula $f'$.
We implement this operation as a deterministic operator with input $f$, $f^L$, $e$, and $\mathrm{pos}(f^L, f)$.

\section{Experiments and results}
% \begin{itemize}
%     \item NDR
%     \item GPT-4
%     \item prompting method
%     \item ours
%     \item analysis of errors (motivation for dynamic windowing)
% \end{itemize}

\begin{figure*}[ht]
    \centering
    \includegraphics[width=0.7\textwidth]{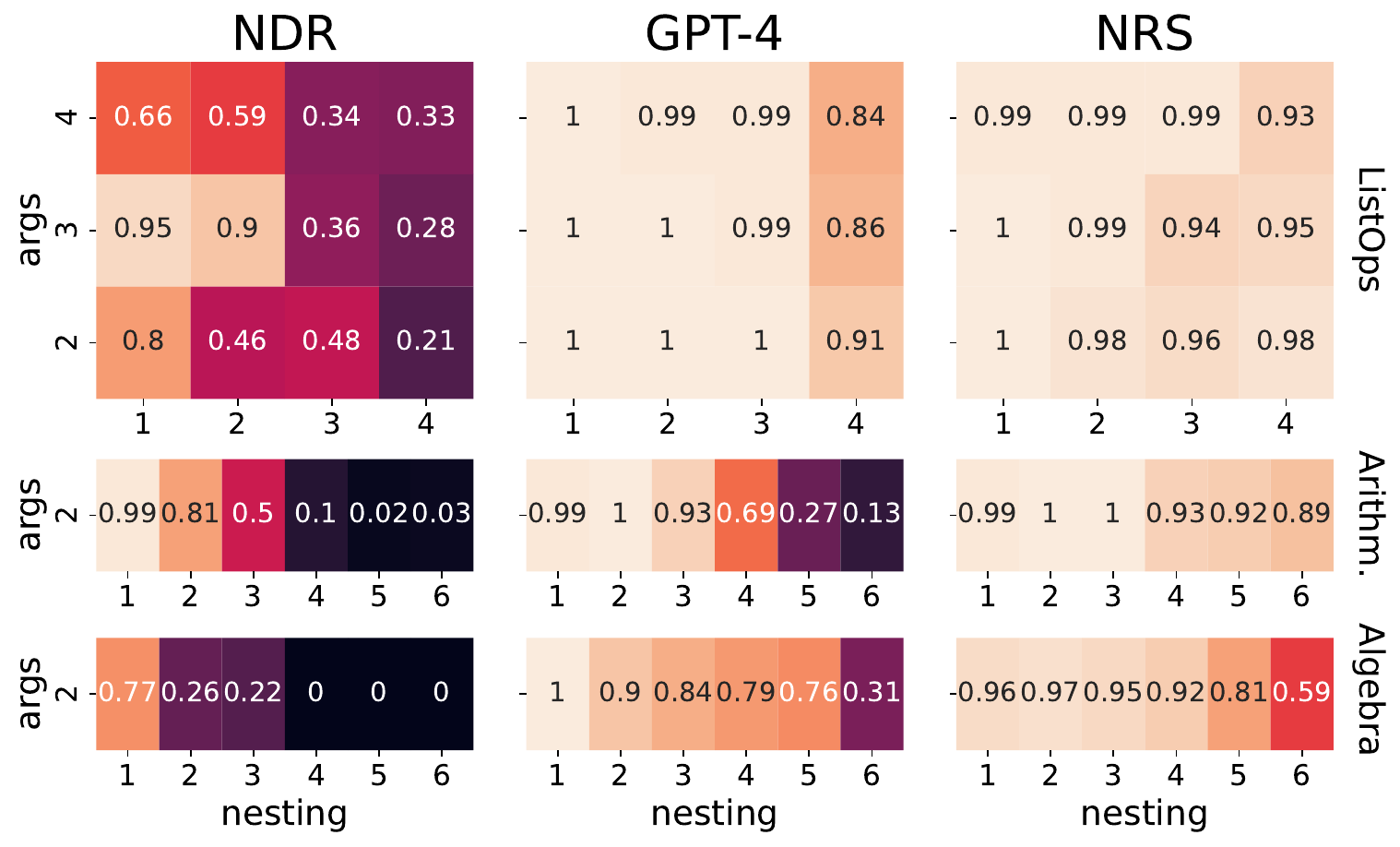}
    
    \caption{Performance of the Neural Data Router, GPT-4 and Neural Rewriting System on ListOps, arithmetic and algebraic formulas.\vspace{0.8cm}}
    \label{fig:acc-tables}
\end{figure*}

We evaluated the capacity of our model to solve formula simplification problems on ListOps, arithmetic, and algebraic formulas.
We compare the performance of the Neural Rewriting System to the Neural Data Router and GPT-4, one of the most advanced LLMs currently available.
\textcolor{black}{Code and data used in our experiments \cite{Petruzzellis2024-mi} and a Technical Appendix containing methodological details and additional results \cite{petruzzellis2024neuralrewritingsolvealgorithmic} are available.}

\subsection{Datasets}
For all three tasks, the automatic generation of formulas is parameterized by specifying the nesting level of a formula.
Any formula is nested at each level in two points: exactly two arguments in the formulas on that level will be other formulas. \textcolor{black}{Formulas are tokenized at the character level for all tasks and trained models.}

{\bf ListOps} The ListOps dataset \cite{DBLP:conf/naacl/NangiaB18} was introduced to evaluate the capacity of neural networks to build parse trees of nested formulas. 
The original dataset consists of formulas composed of operations on lists of integers, including minimum, maximum, median and sum modulo $10$ of a list of integers.
We modified the ListOps dataset so that formulas had exactly 2 nesting points at each nesting level, and we made it possible to specify the number of arguments that appear in formulas at any level.
\textcolor{black}{Since we are interested in the system's capacity to generalize on highly nested formulas rather than learning specific operations}, we reduced the set of operations to minimum, maximum and sum modulo $10$.

{\bf Arithmetic} We generated formulas composed by sum, subtraction and multiplication operations between two integers sampled in the interval $[-99, 99]$.
Since in this work we do not investigate the capacity to generalize to numbers with more digits than those seen during training, we used the modulo $100$ of the intermediate results in the solution process.

{\bf Algebra} We considered a subset of algebraic formulas that can always be reduced up to a minimal form in a deterministic way.
Such formulas involve sum and subtractions between two monomials and their final value is always a monomial.
The numerical coefficients of monomials were sampled in the interval $[-99,99]$, and each monomial can contain up to four literal variables sampled from the set $\{a,b,x,y\}$.
All monomials appearing in a given formula have the same literal variables.
As in the case of Arithmetic, all intermediate numerical values were taken modulo $100$ when computing the final value of the formula.

\subsection{Experiments}
\subsubsection{Neural Rewriting System}
\label{exp-nrs}
We describe here how we built the training and validation sets for the Selector and Solver modules. Further methodological details can be found in the Technical Appendix.

\noindent\textbf{Selector} 
%Indeed, we aim to learn a selection mechanism that can be applied also to more complex formulas.
In Arithmetic and Algebra problems we included in the training set formulas with nesting levels 1, 2 and 3.
%, together with formulas that represent intermediate steps in the resolution process of these formulas.
In the ListOps problem, in which we could specify the number of arguments that appear in the formulas, we included in the training set formulas with nesting levels 1 and 2 that have 2 or 3 arguments for each operator.
%, together with formulas appearing in their resolution process.
In all three problems, the training set also contains atomic elements corresponding to the initial formula's final value. 
Solving formula simplification problems by iteratively simplifying leaf formulas produces several simplifications of the initial formula.
To show to the Selector the complete solution process of formulas in the training set, we have also included in it formulas that appear as intermediate solution steps of the formulas described above. 
%During training, we generated batches containing samples with different structural complexity with equal probability.

We constructed a separate in-distribution validation set with samples with the same structural characteristics as those in the training set.
Unlike common machine learning tasks where models are tested on the same data distribution they are trained on, we want the Selector to possess out-of-distribution (OOD) generalization capabilities to identify the leaf formulas even in longer inputs than those seen during training.
For this reason, we also built a separate out-of-distribution validation set containing formulas with greater structural complexity, \textcolor{black}{and use this set for model selection}.
For Arithmetic and Algebra, we included samples with nesting levels 4, 5 and 6.
%, together with formulas that appear in their resolution process.
For ListOps, we included formulas with nesting levels 3 and 4 with 2, 3 or 4 arguments for each operator.
To select the most capable model across the iterative resolution process, we also included formulas representing examples of intermediate resolution steps.
%, together with formulas that appear in their resolution process.
To control the structural complexity of the formulas, the OOD validation sets are balanced across the nesting level of the leaf expressions.\footnote{In the case of ListOps, we also balance across the number of arguments in the leaf expression.}

\noindent\textbf{Solver} The training and validation set of the Solver included two kinds of samples: leaf formulas, mapped to equivalent atomic elements, and atomic elements themselves, mapped to the end-of-computation special symbol $\omega$. 
During training, we generated batches that included both kinds of samples with equal probability \textcolor{black}{to avoid biasing the model toward solving leaf formulas}.

\subsubsection{Neural Data Router}
\label{exp-ndr}
The Neural Data Router (NDR) is a variant of a transformer encoder designed to learn to solve algorithmic problems with strong out-of-distribution generalization capacity. 
In terms of the formalization of formula simplification problems we have defined in Section \ref{sec:fsp}, the model takes as input a formula $f$ and gives as output the equivalent atomic element $e$.
The NDR thus implements the composition of the $sel$, $sol$, and $com$ functions in one or more transformer layers, assuming that the effective integration between the three happens sub-symbolically in the embedding space during training.

This model has been previously evaluated on relatively simple algorithmic benchmarks that are nevertheless significantly similar to the problems we consider, including the solution of formulas in the original ListOps dataset and simple arithmetic formulas on single-digit integers. The main difference with the problems we consider here is that the operands in our Arithmetic and Algebra problems are of greater complexity. Moreover, in the ListOps problems, we aim to generalize both on the number of operands and on the nesting depth of the formulas, whereas, in the original work, the number of operands observed at training time was the same as that observed at test time.
Furthermore, we use significantly fewer samples during training and of lower complexity, as the NDR was originally trained on arithmetic and ListOps formulas with up to 5 nested operations.

We applied a minor modification to the architecture to adapt the model to the problems we have considered.
Indeed, the problems considered in the original work always had a single-digit integer as a result, which the model was trained to output as the first token in the sequence given as output by the encoder.
Since, in general, this no longer applies to our problems, we read the final answer from the first $k$ positions of the sequence produced by the encoder, where $k$ is the maximum length of a problem's targets.

% It is relevant to notice that the NDR has already been shown to be limited in the solution of algorithmic problems in another work by the same authors where the model was found to have limited ability to apply a known function in a different context than seen during training \cite{DBLP:conf/emnlp/CsordasIS22}.

We build all development sets for the NDR using the same top-level formulas we included in the analogous sets for the Selector. 
Since the NDR was designed to compute intermediate results sub-symbolically, we use as training target the final simplified form of the input formula and we do not include intermediate formulas as we do with the Neural Rewriting System.
\footnote{\textcolor{black}{For comparison, we include in the Technical Appendix results for models trained on larger datasets that also include intermediate formulas.}}
Following the original experimental protocol, we made the training set balanced across nesting levels and number of operands.
As done when training the Selector module, we create in- and out-of-distribution validation sets \textcolor{black}{and use the latter} to optimize hyperparameters, using the Weights and Biases Bayesian search on the same hyperparameters and intervals described in the original work.
We report the final hyperparameter values for each task in the Technical Appendix.

\subsubsection{GPT-4}
Chain-of-Though (CoT) prompting improves the performance of LLMs on reasoning tasks by enabling step-by-step solution procedures.
In terms of the formalization outlined in Section \ref{sec:fsp}, we may observe how such reasoning chains resemble the iterative solution method of nested formulas based on rewriting.
Therefore, we can say that LLMs prompted using CoT-like methods loosely implement the composition of $sel$, $sol$, and $com$ functions in the whole model, where the iterative application of the function is implemented via the auto-regressive generation of outputs.\footnote{Notice that in general, at any given point of the solution process, the context given as input of the model will contain several different simplifications of the input function $f$ rather than just one as it happens with the other models we consider.}

We choose to prompt GPT-4 using self-consistency prompting \cite{DBLP:conf/iclr/0002WSLCNCZ23} combined with zero-shot Chain-of-Thought (CoT)  \cite{DBLP:conf/nips/KojimaGRMI22}.
Zero-shot CoT has been introduced as a simpler alternative to CoT prompting, which allows users to achieve similar performance on reasoning benchmarks without the need to engineer exemplars for few-shot reasoning.
This is achieved by simply making the model's answer start with the sentence: ``Let's think step-by-step''.
After the model has generated a response, it is prompted again to retrieve a well-formatted output.
Self-consistency prompting operates on the concept that reasoning problems can have multiple valid paths leading to the same conclusion. 
To harness this, we generate 10 outputs for each input, allowing us to select the most consistently produced one. 
This method boosts confidence in the model's output and significantly enhances accuracy, resulting in a notable improvement in performance.

The actual prompt used for zero-shot CoT was built giving the model a minimal description of the problem at hand and then asking to solve it.
For example, the zero-shot CoT prompt corresponding to the ListOps input sample \texttt{[MIN[SM54][MIN39]]} is: ``\textit{\texttt{MIN}, \texttt{MAX} and \texttt{SM} are operators on lists of single-digit integers which have the semantics of minimum, maximum and sum modulo 10, respectively. Solve the following expression involving these operators: \texttt{[MIN [SM 5 4] [MIN 3 9]]}.}''.
Following the zero-shot CoT prompting technique, the model was then prompted a second time to extract the well-formatted final answer.

\subsection{Results}
We report in Figure \ref{fig:acc-tables} the performances of all considered models.
All models are tested on 100 samples for each level of complexity.
We measure the sequence accuracy of the models, that is the exact match between the model's output and the target sequence.
When we evaluate GPT-4 on the Algebra task, we take into account equivalent forms of the symbolic outputs using the SymPy library \cite{10.7717/peerj-cs.103}.

We first observe that all the models perform better on simple data splits and tend to worsen on more complex ones.
This indicates that increasing the number of arguments and the nesting of formulas makes the problems more difficult for all models, as expected.

Analyzing the performance of the Neural Data Router, we find that the models with the best hyperparameters configurations have lower accuracy than other models on the simple splits and are not able to generalize the learned solution process to complex problem instances.
Learning to solve simple formulas with a single nesting level -- i.e., learning the $sol$ function -- appears to be the easiest task for the model, although it is learned with almost perfect accuracy only on arithmetic formulas.
The lack of generalization capacity on complex formulas may, therefore, be determined by two different situations: making one or more errors while solving many leaf formulas, or learning a composition of the $sel$, $sol$, and $com$ functions that cannot generalize effectively on nested formulas.
Since the model is unable to generalize both on arithmetic and algebraic formulas, whose leaf formulas can be solved with substantially different levels of accuracy, we argue that both factors could play a role in causing the limited generalization capacity of the model.
% specially on the ones in the arithmetic and algebraic domain, possibly because their leaf formulas contain symbols or highly nonlinear mathematical operations, such as multiplication between two-digit integers.

We observe that GPT-4 can achieve perfect or almost perfect results on leaf formulas and simpler problem instances, surpassing the performance of the Neural Data Router and, in some cases, our architecture.
However, the performance of GPT-4 is significantly lower on deeply nested formulas.
Indeed, while self-consistency prompting improved the model's performance compared to vanilla zero-shot CoT prompting\footnote{We report the performance of GPT-4 with vanilla zero-shot CoT prompting in the Technical Appendix for comparison.}, it has a limited impact on the capacity of the LLM to solve complex formulas, consistently with previous research \cite{lrec-coling24}.
%Comparing the performance achieved on ListOps to that on the other two tasks, we can conjecture that solving deeply nested formulas is harder for GPT-4 than solving formulas with many operands.
%Also in the case of GPT-4, it seems that accurately computing many arithmetic operations between double-digit integers, including multiplication modulo 100, is the hardest task for the LLM.
%It is interesting to notice that, in the case of GPT-4, despite the fact that the solution of individual leaf formulas is almost always exact, the solution of nested formulas involving many such simple operations can be very challenging for the model.

Considering the model as an approximation of the composition of the $sel$, $sol$, and $com$ functions, its limited generalization capability seems to be determined by the poor quality of this approximation obtained with the prompting method we used.
%the difficulty of obtaining an effective simulation of the iterative application of the composite function with the prompting method we considered.
However, by comparing the performance of the model on ListOps formulas with the other two tasks, we observe that the complexity of the individual leaf formulas (much simpler in the case of ListOps) seems to influence the capability of the model to apply a solution method that can generalize systematically on complex formulas.

\begin{figure}[t]
    \centering
    \includegraphics[width=\linewidth]{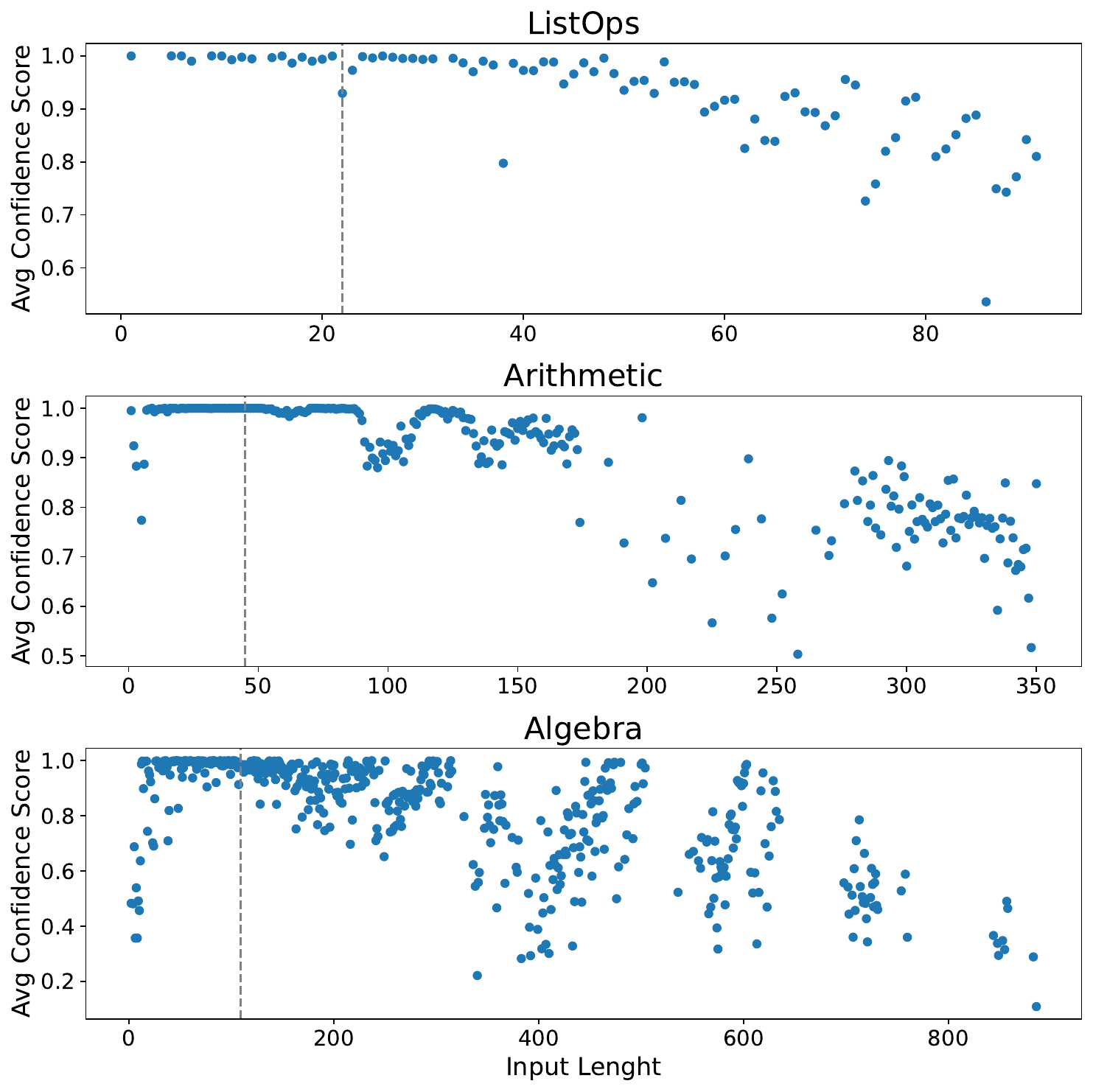}
    \caption{Input length against the average confidence score of 1,000 outputs of the Selector. The vertical lines represent the maximum input length in the training set.}
    \label{fig:conf-scores}
    \vspace{0.8cm}
\end{figure}

% We report both the evaluation of the NRS generating 1000 outputs per input with and without Dynamic Windowing in the Selector.

Coherently with the model selection process, in Figure \ref{fig:acc-tables}, we report the performance of our Neural Rewriting System with 1,000-way multi-output generation, and value of the Dynamic Windowing threshold $T$ of 60, 150 and 200 for the ListOps, Arithmetic and Algebra problems, respectively.
The Neural Rewriting System consistently outperforms the baselines on the three formula simplification tasks we consider.
Importantly, the system shows a significantly high degree of generalization on formulas that are much more complex than those observed during training.
Given the modular nature of our architecture, in the following paragraphs, we will analyze the role of each architecture component in achieving this result.

\begin{figure}[t]
    \centering
    \includegraphics[width=\linewidth]{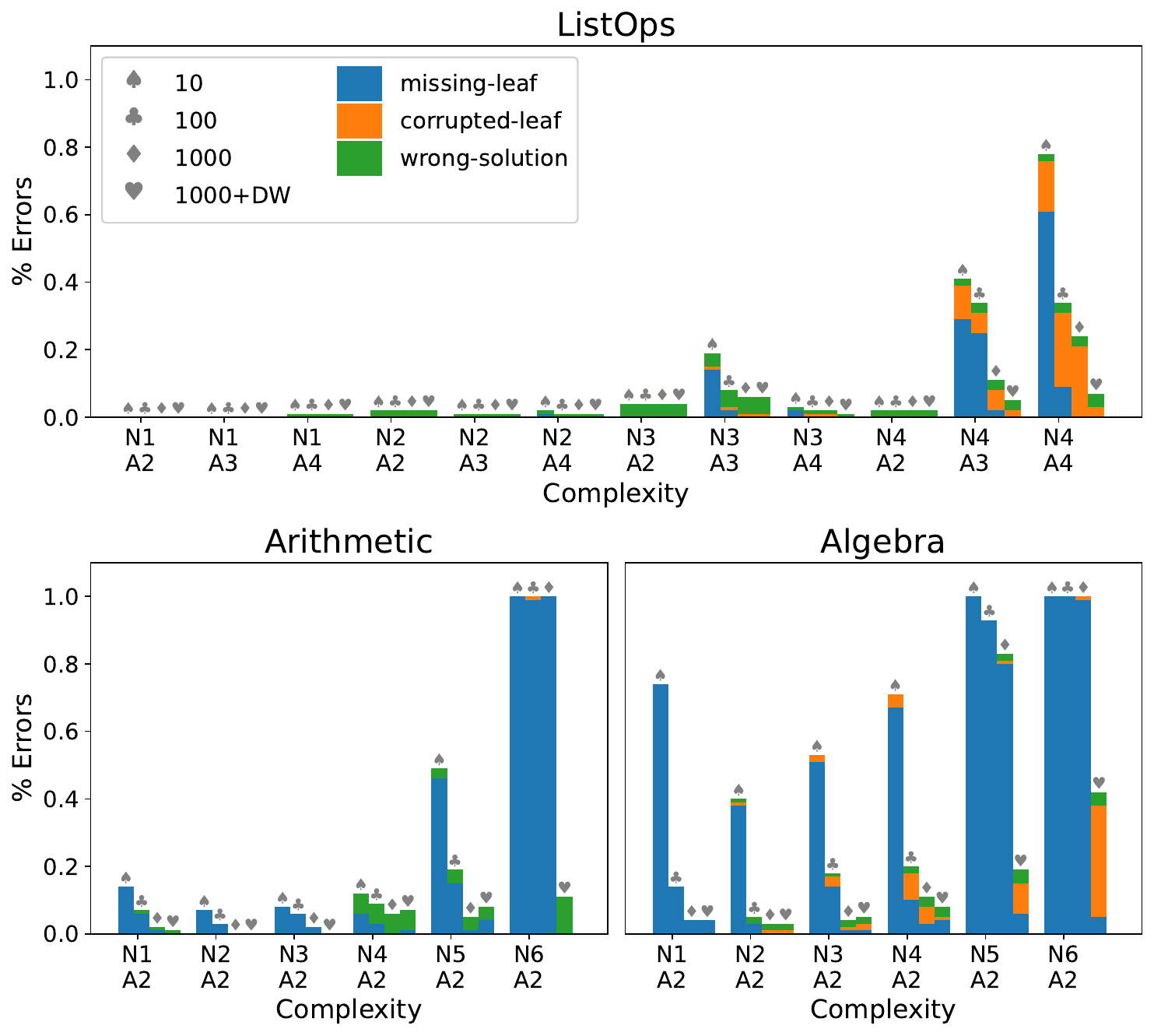}
    \caption{Errors committed by the Neural Rewriting System classified by cause of error. Error bars corresponding to a complexity split refer to models where the Selector generates 10, 100 or 1,000 outputs per input and (optionally) employs the Dynamic Windowing mechanism. Data splits are defined by their nesting level N and number of arguments A.}
    \label{fig:err-analyis}
    \vspace{0.8cm}
\end{figure}

We first analyze the variation of the confidence score corresponding to variations in the length of the input sequences in the test set.
We do so because the confidence score plays an important role in selecting the best Selector output in multi-output generation.
Figure \ref{fig:conf-scores} represents input lengths versus the average confidence score of 1,000 Selector outputs.
The vertical lines in the plot represent the maximum input length seen during training.
While the average confidence score remains close to the training range for outputs that are moderately longer than the longest training samples, it drops significantly for sequences much longer than that.
This can directly determine a decrease in model performance on very long formulas since Selector outputs with a low confidence score are more likely to be incorrect.
Indeed, we set the value of the Dynamic Windowing threshold $T$, which determines the sequences on which the mechanism will be applied, by looking at the points where Selector confidence drops in each task.
We also notice that confidence scores are lower for very short arithmetic and algebraic formulas.
We suppose this is due to the fact that the number of formulas of a given nesting level grows exponentially with the value of the nesting level.
Therefore, the Selector observed fewer examples of formulas with nesting level 1 during training.
However, this does not hinder performance when the Selector generates many outputs per input.

% The dynamic windowing mechanism effectively addresses the drop of confidence on longer sequences.
% We report in Figure \ref{fig:acc-tables} the accuracy of a Neural Rewriting System in which the Selector generates 1000 outputs per input and applies the dynamic windowing mechanism on the longest sequences.
% We selected the length threshold $T$ that is used to specify which sequences will be processed using dynamic windowing based on the analysis of the confidence scores of the Selector discussed above.
% Specifically, we set the thresholds to 60, 150 and 200 for the ListOps, Arithmetic, and Algebra problems, respectively.
 
We finally assessed the impact of the multi-output generation strategy on the model's performance by comparing the errors committed by the model when the Selector generated 10, 100 and 1,000 outputs per input.
We measure the errors separately on subsets of formulas with different nesting levels and number of operands, to understand the impact of increasing the number of outputs generated by the model on the solution of both simple and complex formulas.
Furthermore, for each model configuration and each subset of samples, we analyzed the root cause of the errors committed by the NRS.
We classify the cause of the errors into three categories: errors caused by the Selector when it outputs a leaf formula that is not present in the input formula (\texttt{missing-leaf}); errors caused by the Selector when it outputs a leaf formula that is present in the input formula but does not correspond to a well-formed expression and whose substitution thus causes the corruption of the input formula (\texttt{corrupted-leaf}); and errors caused by the Solver when it outputs a wrong solution for the leaf formula identified by the Selector (\texttt{wrong-solution}).
Figure \ref{fig:err-analyis} represents this multi-faceted measurement of the errors committed by the models.

Looking at the frequency of each of these classes of errors, we clearly see that generating only 10 outputs per input with the Selector is particularly ineffective on algebraic formulas, and on deeply nested arithmetic and ListOps formulas.
Increasing the number of generated outputs to 100 or 1,000 considerably improves the System's performance on most formulas of all three tasks, apart from the most deeply nested arithmetic and algebraic formulas.
On such samples, applying Dynamic Windowing drastically reduces the number of errors the model commits while also bringing further minor improvements on simpler formulas.

As we described in Section \ref{sec:nrs}, the modules in the Neural Rewriting System have been designed as neural implementations of the $sel$, $sol$ and $com$ functions that appear in the simple solution algorithm for formula simplification problems described in Section \ref{sec:fsp}.
For this reason, we can analyze in a greater level of detail to which modules are most error due, and thus which of the functions listed above have been more difficult to reproduce in a neural network.

Given our design choice to implement the $com$ function using a CNN dynamically parameterized by the Selector, the errors due to this module can be directly traced back to errors made by the Selector.
Looking at the classification of errors 
%made by different model configurations on different types of formulas 
represented in Figure \ref{fig:err-analyis}, we can see how the errors due to a \texttt{wrong-solution} of a leaf formula by the Solver are the minority.
Therefore, the $sol$ function appears to be relatively easy to implement in a transformer, consistent with our prior observations on NDR and GPT-4 models.
Errors due to the Selector are instead the vast majority and are only partially mitigated by the generation of multiple outputs and the Dynamic Windowing, especially on algebraic formulas.
The $sel$ function, which formalizes the ability to identify leaf expressions, appears to be the hardest one to model with a transformer-based architecture, particularly due to the necessity to generalize on very long formulas.

\section{Discussion}
In this work, we have considered formula simplification problems, a kind of benchmark used to study the systematic generalization capabilities of neural networks.
We proposed a formal description of this class of problems and used it to define a simple general solution algorithm involving three elementary functions.
We then implemented a general neural architecture called the Neural Rewriting System, which models such algorithm by adopting a modular design.

We have compared the performance of the Neural Rewriting System to that of a highly specialized neural architecture, the Neural Data Router, and of GPT-4, a powerful general-purpose large language model. 
We have tested the models on nested arithmetic and algebraic formulas, and formulas from the ListOps dataset.
We showed that while both the Neural Data Router and GPT-4 are, to different extents, able to solve simple formulas, both models struggle or fail when dealing with deeply nested formulas.
The Neural Rewriting System achieves a considerably greater degree of generalization on much more complex formulas than those it was trained on.
We analyzed models' failures in light of our formalization of formula simplification problems, highlighting the features of these problems that could represent the main obstacles to systematic generalization.

\textbf{Limitations} While we consider formula simplification problems with different number and kinds of operands, we also pose several constraints on these formulas.
For instance, we limit our study to formulas with only one final form which can always be reached independently of the substitution applied.
Both of these conditions are not satisfied, in general, by algebraic expressions, whose form could be made arbitrarily complex and not further reducible.
Future work should thus study the systematic generalization capabilities of neural models in the more general case of formulas with several possible final forms, reachable from different solution paths.
% Furthermore, the architectural mechanisms that we employ to generalize on formula simplification problems rely on the exploitation of specific assumptions that apply to the class of problems consider in this study.
% More general architectural elements and/or training methods should thus be devised to achieve more robust and widely-applicable generalization in neural systems.

\section{Acknowledgements}
We thank OpenAI for granting research access to the GPT APIs.

%%%%%%%%%%%%%%%%%%%%%%%%%%%%%%%%%%%%%%%%%%%%%%%%%%%%%%%%%%%%%%%%%%%%%%%% 

%%%%%%%%%%%%%%%%%%%%%%%%%%%%%%%%%%%%%%%%%%%%%%%%%%%%%%%%%%%%%%%%%%%%%%%%

%%% Use this environment to include acknowledgements (optional).
%%% This will be omitted in doubleblind mode.

% \begin{ack}
% By using the \texttt{ack} environment to insert your (optional) acknowledgements, you can ensure that the text is suppressed whenever you use the texttt{doubleblind} option. In the final version, acknowledgements may be included on the extra page intended for references.
% \end{ack}

%%%%%%%%%%%%%%%%%%%%%%%%%%%%%%%%%%%%%%%%%%%%%%%%%%%%%%%%%%%%%%%%%%%%%%%%

%%% Use this command to include your bibliography file.

\bibliography{bibliography}

\clearpage

\section{Technical Appendix}
\subsection{Model selection and training procedures}
\subsubsection{Neural Rewriting System}
{\bf Solver} We tuned the hyperparameters of the Solver using a random search on the embedding size, the number of encoder and decoder layers, the dropout rate, and the learning rate.
In all models, we used four attention heads and a hidden state in the feed-forward layers that was four times larger than the embedding size.
We trained the models using the Adam optimizer with default parameters, a batch size of 512 and a cosine annealing schedule of the learning rate with a warm-up period of 1500 iterations.
For all tasks, we searched hyperparameters values in the following ranges: \{64, 128, 256\} for the embedding size, [1, 4] for the number of encoder and decoder layers, [0.1, 0.4] for the dropout probability, [1e-5, 1e-4] for the learning rate.
% emb 64, 128, 256
% enc dec layers 1, 2, 3, 4
% dropout 0.1, 0.4
% lr 1e-5, 1e-4

For the ListOps and Algebra tasks we selected transformers with 2 encoder and 2 decoder layers, while for the Arithmetic task we selected transformers with 3 encoder and decoder layers.
For the ListOps task, we trained the models for 10,000 iterations and selected a model with embedding size 128, trained with dropout probability 0.33 and learning rate 5e-5.
For the Arithmetic task, we trained the models for 100,000 iterations and selected a model with embedding size 256, trained with dropout probability 0.1 and learning rate 9e-5.
For the Algebra task, we trained the models for 40,000 iterations and selected a model with embedding size 256, trained with dropout probability 0.33 and learning rate 8e-5.

{\bf Selector} As for the Solver, in all models we used four attention heads and a hidden state in the feed-forward layers that was four times larger than the embedding size.
We trained the models using the Adam optimizer with default parameters, a batch size of 512 (256 for Algebra) and a cosine annealing schedule of the learning rate with warm-up.
We tuned the embedding size, the number of encoder and decoder layers, the width of the diagonal window applied to the self-attention matrix, the dropout rate, the learning rate, the number of warm-up iterations and the value of gain parameter for initialization of the self-attention layers using a random search.
For all tasks, we searched hyperparameters values in the following ranges: \{128, 256, 512\} for the embedding size, [1, 6] for the number of encoder and decoder layers, [0.1, 0.4] for the dropout probability, [1e-5, 6e-5] for the learning rate, [500, 3,000] for the number of warm-up iterations, [0.5, 1.5] for the initialization gain parameter.
% specify ranges of values?
% emb [128, 256, 512]
% layers enc/dec [1, 6]
% mask width depends
% drop [0.1, 0.4]
% lr [1e-5, 6e-5]
% warmup [500, 3000]
% mha init [0.5, 1.5]

For the ListOps task, we trained the models for 20,000 iterations and chose a model with embedding size 256, 1 encoder layer and 2 decoder layers, a diagonal window of width 5, self-attention initialization gain of 1.3, trained with dropout probability of 0.15, learning rate of 5e-5 and 1200 warm-up iterations.
% Arithmeitc
% emb 256
% layers enc/dec 4/2
% mask width 6
% drop 0.35
% lr 1e-5
% warmup 2200
% mha init 1.2
For the Arithmetic task, we trained the models for 30,000 iterations and chose a model with embedding size 256, 4 encoder layers and 2 decoder layers, a diagonal window of width 6, self-attention initialization gain of 1.2, trained with dropout probability of 0.35, learning rate of 1e-5 and 2,200 warm-up iterations.
% Algebra
% emb 256
% layers enc/dec 4/2
% mask width 11
% drop 0.3
% lr 4e-5
% warmup 2000
% mha init 1.3
For the Algebra task, we trained the models for 30,000 iterations and chose a model with embedding size 256, 4 encoder layers and 2 decoder layers, a diagonal window of width 11, self-attention initialization gain of 1.3, trained with dropout probability of 0.3, learning rate of 4e-5 and 2,000 warm-up iterations.
\vfill\null

\subsubsection{Neural Data Router}
Here we report the hyperparameters of the best Neural Data Router configuration we selected for each problem.
We searched hyperparameters values in the same ranges used in the original paper.

For the ListOps task, we selected a model with 8 encoder layers, 512-dimensional state, 16 attention heads, 1024-dimensional feed-forward hidden layer, trained with learning rate 5e-4, dropout probability 0.3, attention dropout 0.46 and weight decay 0.06.
All models were trained for 30,000 iterations, with batch size 512 and the AdamW optimizer.
For the Arithmetic task, we selected a model with 7 encoder layers, 512-dimensional state, 8 attention heads, 2048-dimensional feed-forward hidden layer, trained with learning rate 6e-4, dropout probability 0.5, attention dropout 0.08 and weight decay 0.05.
All models were trained for 100,000 iterations, with batch size 512 and the AdamW optimizer.
For the Algebra task, we selected a model with 13 encoder layers, 512-dimensional state, 16 attention heads, 2048-dimensional feed-forward hidden layer, trained with learning rate 4e-4, dropout probability 0.1, attention dropout 0.3 and weight decay 0.02.
All models were trained for 100,000 iterations, with batch size 256 and the AdamW optimizer.

\subsection{Additional results}
\subsubsection{NDR trained on a larger dataset}
\begin{figure}
    \centering
    \includegraphics[width=\linewidth,scale=0.3]{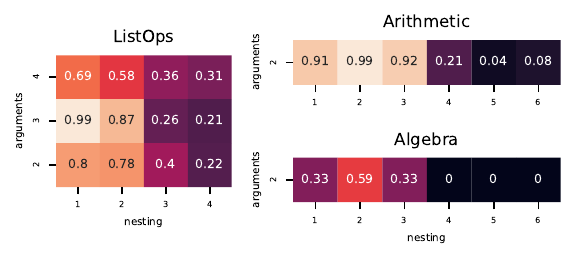}
    \caption{Performance of the NDR trained also on intermediate formulas.}
    \label{fig:acc-ndrexp}
\end{figure}

\textcolor{black}{We report in Figure \ref{fig:acc-ndrexp} the accuracy of the outputs generated by a Neural Data Router trained on a larger dataset equivalent to the one used to train the Selector, i.e., including also intermediate formulas (see Sections \ref{exp-nrs} and \ref{exp-ndr}).}

\subsubsection{GPT-4 zero-shot CoT}
\begin{figure}
    \centering
    \includegraphics[width=\linewidth,scale=0.3]{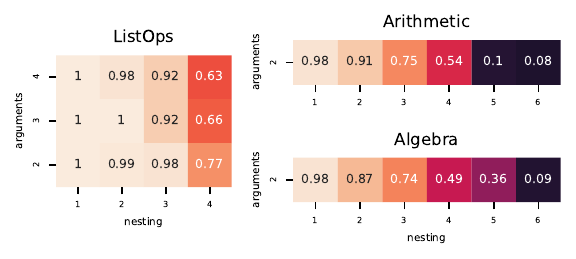}
    \caption{Performance of GPT-4 with zero-shot CoT prompting.}
    \label{fig:acc-zscot}
\end{figure}

We report in Figure \ref{fig:acc-zscot} the accuracy of the outputs generated by GPT-4 probed with vanilla zero-shot Chain-of-Thought prompting.

\end{document}